\documentclass[a4paper]{article}

\usepackage{fullpage} 
\usepackage{parskip} 
\usepackage{tikz} 
\usepackage{amsmath}
\usepackage{hyperref}
\usepackage{subfig}

\usepackage{graphicx}             
\usepackage{moreverb}             

\usepackage{graphicx}
\usepackage{amsmath,amssymb} 
\usepackage{color}
\usepackage{blindtext}
\usepackage{tabto, hyperref}
\usepackage{subfig, array}
\usepackage{blindtext, amsmath}
\usepackage{graphicx, tabto, hyperref}
\usepackage{subfig}
\usepackage{dblfloatfix}
\usepackage{graphics}
\usepackage{fontenc}
\usepackage{txfonts}
\usepackage{pifont}
\usepackage{color}
\usepackage{hyperref}
\usepackage{booktabs}
\usepackage{multirow}
\usepackage{float}
\usepackage{siunitx, array}
\usepackage{authblk}

\title{LOOP Descriptor: Local Optimal Oriented Pattern}

\author[1]{Tapabrata Chakraborti}
\author[1]{Brendan McCane}
\author[1]{Steven Mills}
\author[2]{Umapada Pal}

\affil[1]{Dept. of Computer Science, University of Otago, NZ}
\affil[2]{CVPR Unit, Indian Statistical Institute, India}

\begin{document}

\maketitle

\begin{abstract}

We introduce the LOOP binary descriptor (local optimal oriented pattern) that encodes rotation invariance into the main formulation itself. This makes any post processing stage for rotation invariance redundant and improves on both accuracy and time complexity. We consider fine-grained lepidoptera (moth/butterfly) species recognition as the representative problem since it involves repetition of localized patterns and textures that may be  exploited for discrimination. We evaluate the performance of LOOP against its predecessors as well as few other popular descriptors. Besides experiments on standard benchmarks, we also introduce a new small image dataset on  NZ Lepidoptera. LOOP performs as well or better on all datasets evaluated compared to previous binary descriptors. The new dataset and demo code of the proposed method are available through the lead author's academic webpage and GitHub.

\end{abstract}

\section{Introduction}
Local binary descriptors have been shown to be effective encoders of repeated local patterns for robust discrimination in several visual recognition tasks [1]. He and Wang [2], in their seminal paper on the subject, introduced the concept of textons. These encode localized textures/patterns in an image into binary words, and the frequency histogram of these words describes the image. The first popular implementation was local binary pattern (LBP) by Ojala et al [3]. 

Since then, many interesting modifications and improvements of these descriptors have been developed. A few of these are modified census transform (MCT) [4], local gradient pattern (LGP) [5], local directional pattern (LDP) [6], uniform local binary pattern (ULBP) [17], etc. LBP encodes the local intensity variation in the neighborhood of each image pixel into a binary word, the decimal equivalent of which then acts as a representative feature encapsulating the pattern of local intensity variation in that neighborhood. The histogram of the LBP values over all pixels of an image serves as the descriptor for that image. MCT [4] is a modified version of LBP, where instead of taking the central pixel intensity of the neighborhood as the threshold, the average intensity over the entire neighborhood is treated as the threshold while forming the binary word. LGP [5] considers the gradient of neighboring pixel intensities with respect to the central pixel intensity to generate the binary word. Both MCT and LGP were demonstrated to be more robust to noise than the original LBP operator [5],[6]. LDP was proposed by Jabid et al. [6] as a more noise invariant alternative to LBP. It incorporates the outputs of the $3\times3$ Kirsch masks in 8 directions, for each pixel, into a binary word and its corresponding decimal value is the desired LDP value, corresponding to that pixel.

An inherent shortcoming of both LBP and LDP, as well as many of the other related descriptors like MCT, LGP, etc., is that each assigns an ad hoc bias while sequencing the generated binary digits to form the binary word. This is because the choice of the initial pixel of this circular sequence of bits determines the binary-to-decimal conversion weight assigned to each of the binary digits, as explained later in the paper. This adds an inherent rotation variance. A few rotation invariant versions have also been suggested in existing literature, like LDP$^{ri}$ [16] for LDP. But these have their own set of limitations and the present work presents an alternative rotation invariant scheme.

The main contributions of this paper are:
\begin{enumerate}
    \item \textbf{LOOP descriptor}: We encapsulate rotation invariance into the main formulation of local binary descriptor, thus overcoming a drawback of most existing descriptors of the genre. In the process we decrease post processing time complexity and increase accuracy of classification. The demo code for LOOP is available through the lead author's GitHub account at \emph{https://github.com/tapabrata-chakraborti}.
    \item \textbf{NZ Lepidoptera dataset}: We present a new fine-grained image dataset of NZ Lepidoptera. Small fine-grained datasets can have significant real life applications, like native/endemic species recognition and bio-diversity analysis. It is difficult to effectively train deep networks on such small specialized datasets, and hence fine-tuned approaches like the present one still have practical significance. The dataset is available at the lead author's academic webpage at \emph{https://tapabrata-chakraborti.github.io}.
\end{enumerate}
Note that though we have shown the efficacy of the proposed method on butterfly and moth species recognition, the methodology may be applied to other similar applications exhibiting repeated local patterns/textures.

\section{Previous Descriptors}

A brief review of LBP and LDP is presented here, since the proposed LOOP descriptor is an improvement designed on them.

\subsection{Local Binary Pattern (LBP)}

LBP [3] is a popular descriptor which captures the local intensity variation patterns of an image and has good discrimination characteristics.

Let $i_c$ be the intensity of an image $I$ at pixel ($x_c, y_c$) and $i_n$ ($n=0, \dots ,7$) be the intensity of a pixel in the $3\times3$ neighborhood of ($x_c, y_c$) excluding the center pixel $i_c$.

Then the LBP value for the pixel ($x_c, y_c$) is given by
\begin{equation}
LBP(x_c,y_c)=\sum_{n=0}^{7} s(i_n-i_c).2^n
\end{equation}
where
\begin{equation}
s(x) =  \left\{
  \begin{array}{l l}
    1 & \quad \text{if $x \geq 0$}\\
    0 & \quad \text{otherwise}
  \end{array} \right.\
\end{equation}

A major disadvantage of LBP is the arbitrary sequence of binarization weights. Depending on the chosen starting pixel of the sequence of binary weights ($2^n, n=0,\dots,7$), the 8 neighbors of the output $3\times3$ grid are allocated subsequent weightage $n$ sequentially. There is no clear logic behind the proper assignment of weight and the result obtained is susceptible to rotation variance. The same pattern rotated between images of the same class or even within different parts of the same image will generate a different binary word, thereby confounding the classification process. In fact, this bias has persisted over to other related descriptors as well, like LDP, MCT, LGP, etc.

\subsection{Local Directional Pattern (LDP)}

LDP is an improved local pattern descriptor which incorporates a directional component by using Kirsch compass kernels. It was shown to be less susceptible to noise than the traditional LBP operator [6].

Let $i_c$ be the intensity of an image $I$ at pixel $(x_c, y_c)$ and $i_n$ $, n=0,1,\dots,7$ be the intensity of a pixel in the $3\times3$ neighborhood of $(x_c, y_c)$ excluding the center pixel $i_c$.
$3\times3$ Kirsch edge detectors centered at $(x_c,y_c)$ in eight possible directions are given in Fig. 1.

The 8 responses of the Kirsch masks are $m_n$, $n=0,\dots,7$ corresponding to pixels with intensity $i_n$, $n=0,\dots,7$ and let $m_k$ be the $k^{th}$ highest Kirsch activation. Then all the neighboring pixels having Kirsch response higher than $m_k$ is assigned 1, and others 0.

But the empirically assigned value of $k$ is \emph{ad hoc} [6]. This fixes the possible number of ones to $k-1$ and number of zeros to $(n+1-(k-1)=n-k+2$ where $n$ is as defined above by the neighborhood pixel number. Hence the possible number of binary words is reduced from $2^{(n+1)}$ to $C^{n+1}_{k-1}$.

Then the LDP value for the pixel $(x_c, y_c)$ is given by
\begin{equation}
LDP_k(x_c,y_c)=\sum_{n=0}^{7} s(m_n-m_k).2^n
\end{equation}
where
\begin{equation}
s(x) =  \left\{
  \begin{array}{l l}
    1 & \quad \text{if $x \geq 0$}\\
    0 & \quad \text{otherwise}
  \end{array} \right.\
\end{equation}

\textbf{Rotation Invariant LDP ($LDP^{ri}$)}: A rotation invariant version was introduced in [16]. Here the neighbor pixel with highest Kirsch mask output is assigned the highest order in the binary word, and then the other bits are taken sequentially as in previous formulations. Thus it assigns an emperical rule to the starting point of the binary word construction. However, it suffers from the self-imposed restriction of always having a leading 1, which immediately reduces the number of available combinations in the binary word by half. The problem of fixed number of 1s and 0s also persists from the original LDP, depending on the value of the threshold $k$.

\section{Proposed Methodology}

The proposed Local Optimal Oriented Pattern (LOOP) is described here in details. 

\subsection{Local Optimal Oriented Pattern (LOOP)}

As discussed earlier, the major disadvantage of LBP and LDP is the arbitrary sequence of binarization weights that adds dependancy to orientation. LDP also suffers from the empirical assignment of value to the threshold variable, which puts an \emph{ad hoc} restriction on the number of bits allowed to be 1, thus reducing the number of possible words, as discussed before. LOOP presents a non-linear amalgamation of LBP and LDP that overcomes these drawbacks while preserving the strengths of each. 

Let $i_c$ be the intensity of an image $I$ at pixel $(x_c, y_c)$ and $i_n$ $(n=0,1,\dots,7)$ be the intensity of a pixel in the $3\times3$ neighborhood of $(x_c, y_c)$ excluding the center pixel $i_c$. The 8 Kirsch masks, as used in LDP previously, are oriented in the direction of these 8 neighboring pixels $i_n$ $(n=0,1,\dots,7)$ thus giving a measure of the strength of intensity variation in those directions, respectively.

This leads us to propose the incorporation of this information by assigning the binarization weight to each neighboring pixel corresponding to the strength of Kirsch output in the direction of that pixel. The underlying rationale behind this approach is that the Kirsch mask output in a particular direction provides an indication of the probability of occurrence of an edge in that direction. Since the LBP indicates the intensity variation over the neighboring pixels in the same directions, the value of the Kirsch output is employed to assign the decimal-to-binary weightage.

As discussed earlier, the 8 responses of the Kirsch masks are $m_n$ corresponding to pixels with intensity $i_n$, $n=0,\dots,7$ 
Each of these pixels are assigned an exponential $w_n$ (a digit between 0 and 7) according to the rank of the magnitude of $m_n$ among the 8 Kirsch mask outputs.

Then the LOOP value for the pixel $(x_c, y_c)$ is given by
\begin{equation}
LOOP(x_c,y_c)=\sum_{n=0}^{7} s(i_n-i_c).2^{w_n}
\end{equation}
where
\begin{equation}
s(x) =  \left\{
  \begin{array}{l l}
    1 & \quad \text{if $x \geq 0$}\\
    0 & \quad \text{otherwise}
  \end{array} \right.\
\end{equation}

Thus the LOOP descriptor encodes rotation invariance into the main formulation. Moreover, the proposed LOOP algorithm also negates the empirical assignment of the value of the parameter $k$ in the traditional LDP method (eqn. 3).

\subsection{Scale and Rotation Invariance}

A multi-scaled amalgamated histogram is constructed to achieve scale-independence. This is done by forming a spatial gaussian pyramid and then concatenating the histograms of LOOP values obtained at each scale to form the final histogram which acts as the descriptor for the image.

Figure 1 illustrates the rotation invariance property of LOOP descriptor, compared to the lack thereof in LBP and LDP. As demonstrated in Figure 1, binary words are formed according to LBP rule and the weights are assigned according to the LDP mask activations. 

\textbf{Tie break in weight assignment:} Referring to Figure 1, the one with more differing nearest neighbour is assigned higher weight. Eg. For the two -2155 for pattern 2, the nearest neighbours for one are -2155 and -275 (difference is 1880) while nearest neighbours for the other are -2155 and -2035 (difference is 120). So the former is assigned higher weight than latter ($2^1$ vs. $2^0$).

\begin{figure}[!t]
\centering
\includegraphics[scale=.65]{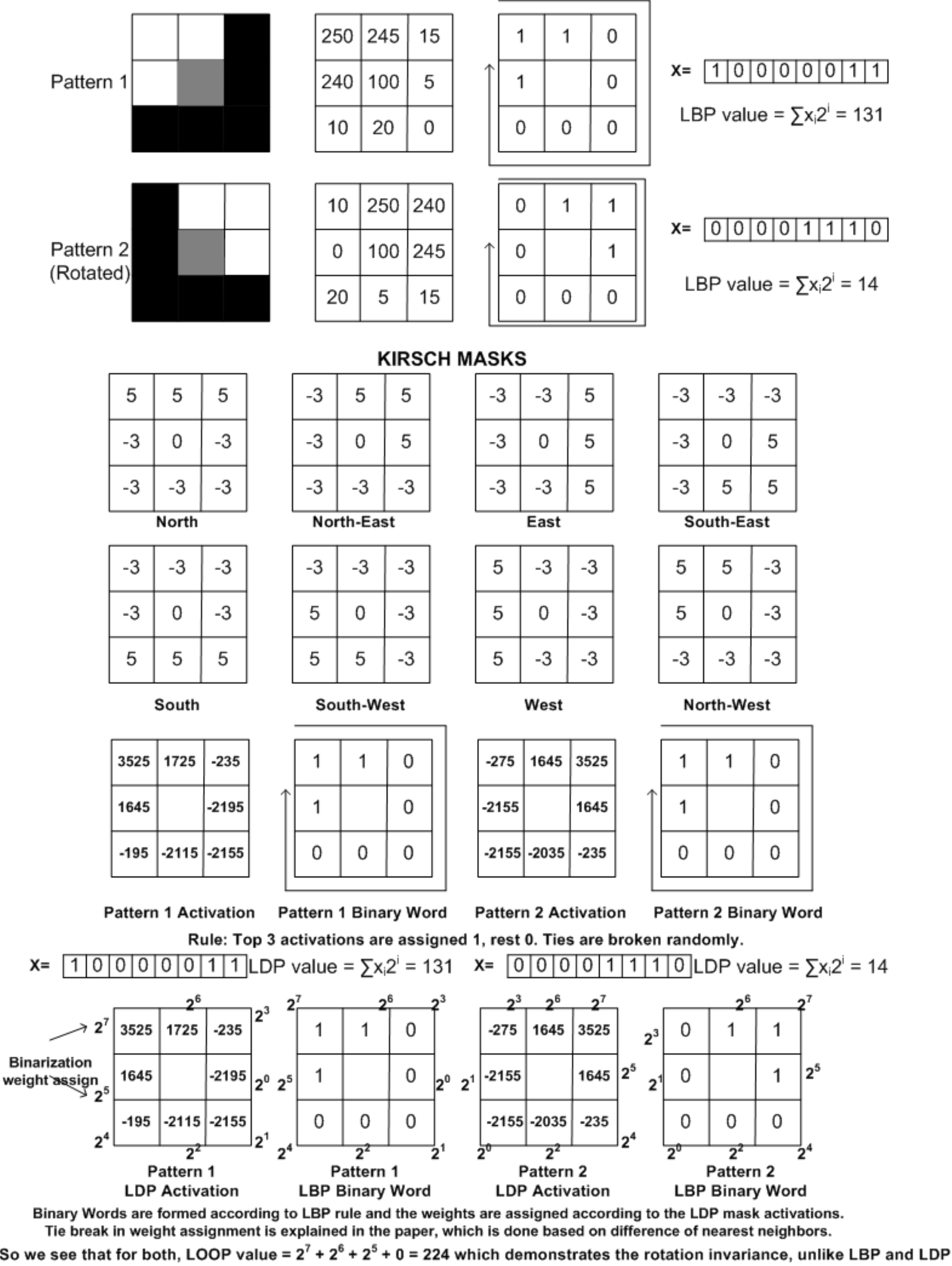}
\caption{Numerical Example to calculate LBP, LDP and LOOP.}
\end{figure}

\section{Experiments and Results}

 \subsection{Experimental Setup}

\textbf{Datasets.} \ 3 image datasets of moth and butterfly species have been used to showcase the performance of the methods.

\emph{Leeds butterfly dataset} [7] was built in 2009 at the University of Leeds, UK. It contains 832 images of 10 species of butterflies with 55 to 100 images per catagory. 

\emph{Ponce butterfly dataset} [8] was built by the Ponce Group at Beckman Institute, University of Illinois at Urbana-Champaign in 2004. It has 7 butterfly types with a total of 619 images. The smallest class contains 42 images and the largest class has 111 images. 

\emph{NZ Lepidoptera dataset} is a new benchmark built during this work at the Department of Computer Science, University of Otago, NZ in collaboration with the CVPR Unit, Indian Statistical Institute. It has 8 classes of NZ butterflies and moths, 4 categories each. Currently it has 640 images with 80 images per class, subject to expansion in near future. Images of NZ moths have been compiled from the publicly available database of NZ Landcare Research. Sample images of the NZ Lepidoptera dataset are presented in Figure 2.\\

\begin{figure}[t!]
\centering

\subfloat[Admiral]{\includegraphics[width=1.3in,height=1in]{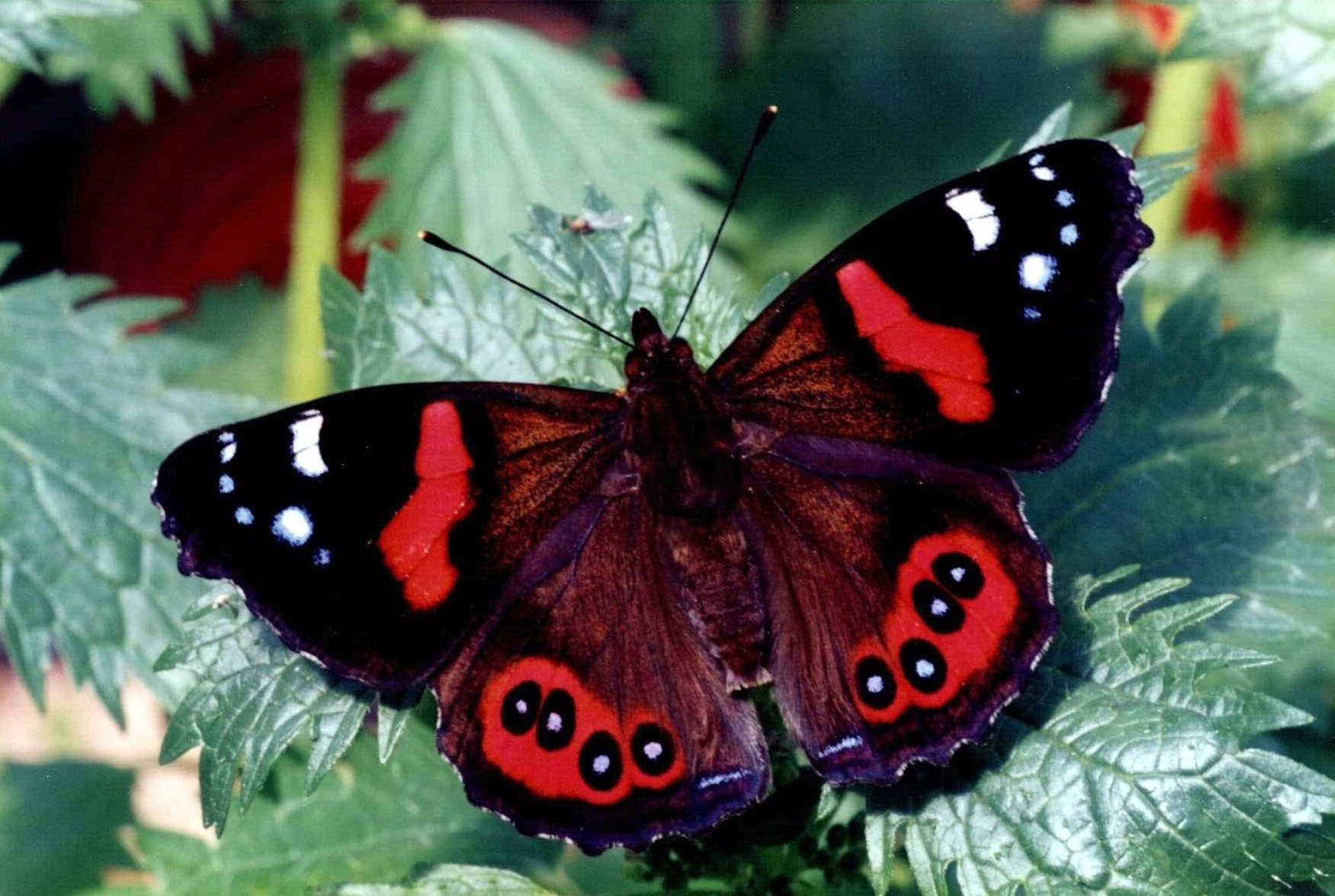}}
\hfil
\subfloat[Blue]{\includegraphics[width=1.3in,height=1in]{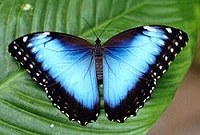}}
\hfil
\subfloat[Copper]{\includegraphics[width=1.3in,height=1in]{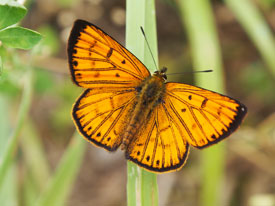}}
\hfil
\subfloat[Ringlet]{\includegraphics[width=1.3in,height=1in]{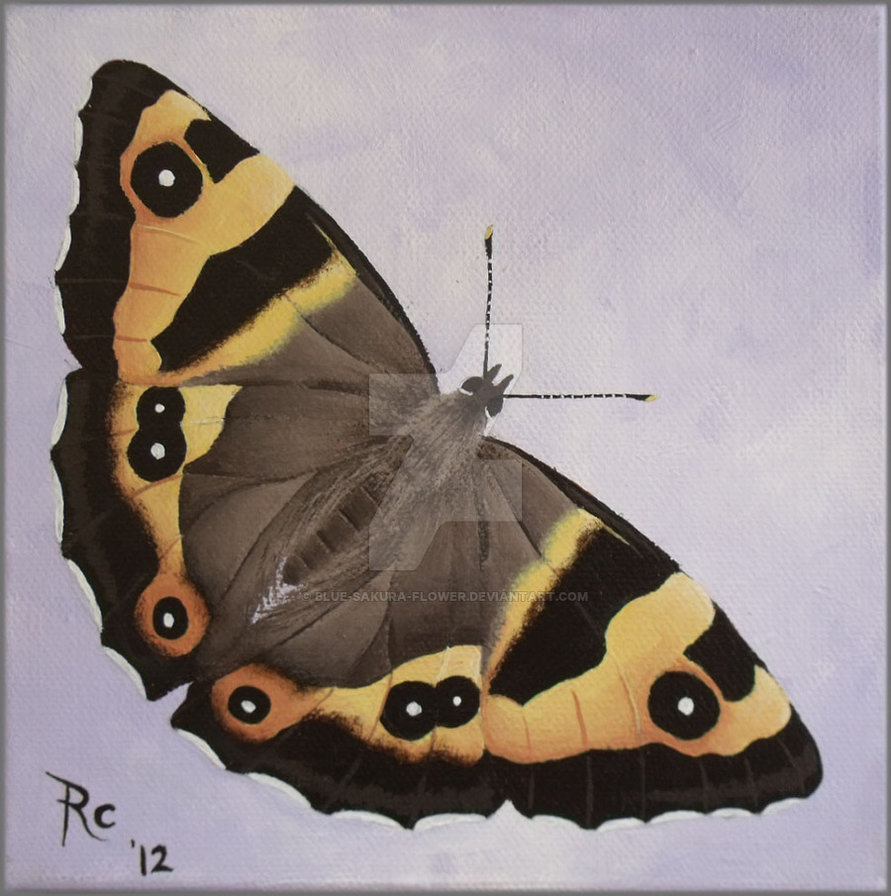}}
\\
\subfloat[Erebidae]{\includegraphics[width=1.3in,height=1in]{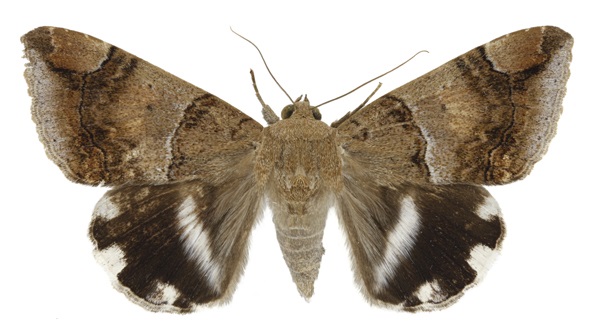}}
\hfil
\subfloat[Geometridae]{\includegraphics[width=1.3in,height=1in]{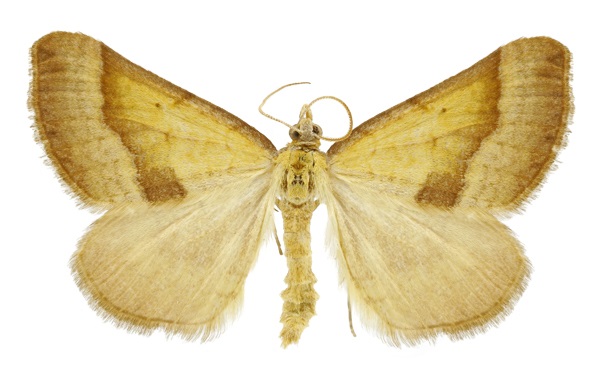}}
\hfil
\subfloat[Hepialidae]{\includegraphics[width=1.3in,height=1in]{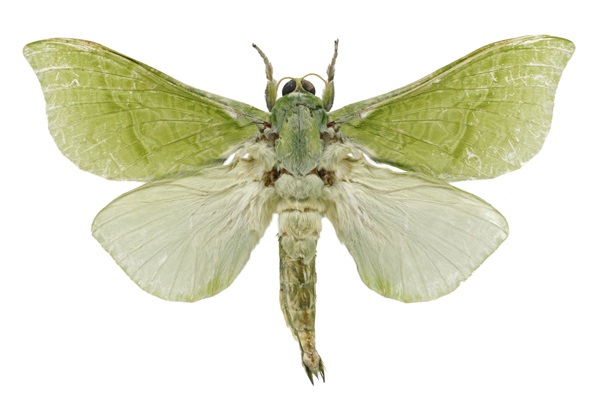}}
\hfil
\subfloat[Noctuidae]{\includegraphics[width=1.3in,height=1in]{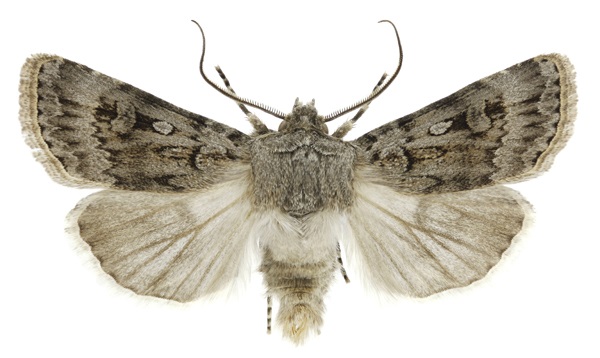}}

\caption{Sample images from each class of the new NZ Lepidoptera dataset. It has 640 images currently with 80 images per class. (a)-(d) are butterfly species \& (e)-(h) are moth species.}
\end{figure}

\textbf{Classifiers.} \ The collaborative representation classifier (CRC) [9] has been shown recently to be effective in handling small fine-grained datasets where the differences in objects between classes is subtle compared to randomized significant background variation within and between classes [10]. CRC represents the test image as an optimal weighted average of training images of all classes and the subsequent residual per class is used to calculate the predicted category. We adopt a recent Probabilistic formulation of CRC called ProCRC [11].

For comparison with a popular off-the-shelf classifier, a support vector machine (SVM) [12] with a $\chi2$ kernel is chosen with settings as in the ProCRC paper [11] for fair comparison. Multiclass categorization is performed with the binary SVM classifier in a one-versus-all fashion in turns. \\

\textbf{Descriptors.} \  We compare the performance of several local binary pattern encoders like LBP, MCT, LDP, LDP$^{ri}$, LGP with the proposed LOOP descriptor. Among these, LOOP is built influenced by LBP and LDP, while MCT and LGP are chosen as relevant modifications of these methods. LDP$^{ri}$ is a rotation invariant modification of LDP. We also compare with three popular modern binary descriptors: BRIEF (Binary Robust Independent Elementary Features) [13], BRISK (Binary Robust Invariant Scalable Keypoints) [14], and ORB (Oriented FAST and rotated BRIEF) [15]. BRIEF was the first of these and presents a simple configuration similar to LBP, without rotation invariance. ORB descriptor is rotation invariant and uses an optimal sampling pair. BRISK has both of these attributes and also has the additional characteristic of a more advanced hand-crafted sampling pattern composed of concentric rings.

\begin{figure}[t!]
\centering

\subfloat[Sample image]{\includegraphics[width=2.7in,height=2.5in]{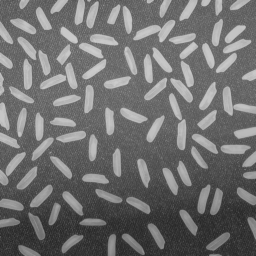}}
\hfil
\subfloat[LOOP output]{\includegraphics[width=2.7in,height=2.5in]{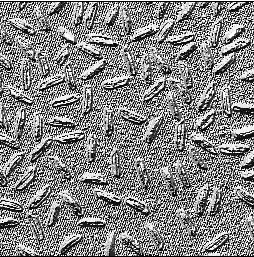}}

\caption{Standard test image (rice.png) and the LOOP output.}
\end{figure}

\subsection{Results and Discussion}

For each dataset, experiments are conducted with 5 fold cross validation and percentage classification accuracies are presented in Table 1 with the highest accuracy in each column highlighted in bold. The LOOP output on a standard test image 'rice.png' is presented in Figure 3 as an illustrative example.

It is observed that LOOP performs much better than LBP and LDP, the two descriptors on which it is based. Performance of LOOP is also better than the competing LBP variants: MCT, LGP and LDP$^{ri}$. LOOP successfully outperforms BRIEF, but has comparable results with BRISK and ORB, both of which are rotation invariant. However, LOOP has a simpler formulation and lower execution time than ORB and BRISK. Using the standard tic-toc functionalities of MATLAB, it is observed that ORB and BRISK have 21\% and 27\% higher computation time respectively than LOOP.

\begin{table}[ht!]
\renewcommand{\arraystretch}{1.5}

\caption{Classification Accuracy (\%)}
\label{table_example}
\centering
\resizebox{0.9\textwidth}{!}{%
\begin{tabular}{|l|c|c|c|c|c|c|}
\hline
&\multicolumn{2}{|c|}{\textbf{Leeds}} & \multicolumn{2}{|c|}{\textbf{Ponce}} & \multicolumn{2}{|c|}{\textbf{NZ}}\\
\hline
& \textbf{SVM} & \textbf{ProCRC} & \textbf{SVM} & \textbf{ProCRC} & \textbf{SVM} & \textbf{ProCRC} \\
\hline
\textbf{LBP} & 62.1  & 64.7  & 68.4  & 70.5  & 55.6 & 59.3 \\
\hline
\textbf{MCT} & 63.9 & 65.8 & 69.9 & 72.3 & 58.7 & 61.1 \\
\hline
\textbf{LDP} & 66.6 & 68.5 & 71.7 & 74.1 & 60.9 & 64.5 \\
\hline
\textbf{LDP$^{ri}$} & 69.2 & 72.5 & 75.1 & 77.9 & 64.8 & 68.3 \\
\hline
\textbf{LGP} & 69.4 & 72.9 & 75.0 & 77.6 & 64.2 & 68.6 \\
\hline
\textbf{BRIEF} & 65.5 & 67.1 & 70.4 & 73.6 & 59.9 & 63.0 \\
\hline
\textbf{BRISK} & 69.8 & 73.5 & 77.7 & 79.3 & \textbf{66.2} & 69.5 \\
\hline
\textbf{ORB} & 71.0 & 73.8 & \textbf{78.4} & 79.9 & 65.8 & 70.1 \\
\hline
\textbf{LOOP} & \textbf{71.5} & \textbf{74.4} & 78.3 & \textbf{80.4} & 66.0 & \textbf{70.6} \\
\hline
\end{tabular}}
\end{table}

Among the competing classifiers listed in Table1, only LDP$^{ri}$, BRISK and ORB are rotation invariant. LOOP yields comparable results to BRISK and ORB, but has lower run time due to simpler formulation. LOOP yields only marginal improvement in performance compared to LDP$^{ri}$, but has similar complexity in formulation and comparable run time. We determine next whether the increase in average accuracy of LOOP over LDP$^{ri}$ is statistically significant. 

\textbf {Sign Binomial Test.} For each descriptor, we have 2 classifiers and 3 datasets, hence 6 combinations per descriptor. Also there are 5 fold cross-validation per combination. So for each descriptor we have 30 sets of accuracy results. Assuming the null hypothesis to be that the two competing methods (LOOP and LDP$^{ri}$) are equally good, then there is 50\% chance of each beating the other. 

It is observed that of the 30 experimental runs, LOOP outperforms LDP$^{ri}$ 22 times. The one-tail P value at 5\% level of significance is 0.0081. Now using Bonferroni correction, at 5\% level of significance, the corrected $\alpha$ for the 6 combinations (2 classifiers and 3 datasets per descriptor) is $0.05/6 =  0.0083$. Since the calculated chance is 0.0081 (less than the corrected $\alpha$), we can reject the null hypothesis and conclude that LOOP has a statistically significant better performance than LDP$^{ri}$.

Also as explained earlier, LDP$^{ri}$ has the constraint of always having a leading 1 thus halving the number of possible words, along with the restriction of having a fixed number of 1s and 0s in the binary word. LOOP is free from these limitations design.

\section{Conclusion}

A novel binary local pattern descriptor, LOOP, which overcomes some disadvantages of its predecessors LBP and LDP, is presented. It is tested on Lepidoptera species recognition with encouraging initial results that warrant further exploration. It outperforms the descriptors on which it is based, along with a few other variants. It has comparable results with popular binary descriptors like BRISK and ORB, but gains in time complexity. 

A new benchmark dataset of NZ Lepidoptera images is also introduced. The proposed method may be used in similar fine-grained applications with small datasets, where fine-tuning a deep network is difficult (limited samples for problems like endangered species recognition, rare pathology detection). 

This letter has only presented experiments and results on one representative problem, that of Lepidoptera classification. But LOOP is a generalized binary descriptor and may be used in further research for other small fine-grained datasets.


\bibliographystyle{plain}
\bibliography{References}

\end{document}